\documentclass[10pt,twocolumn,letterpaper]{article}

\usepackage{cvpr}
\usepackage{times}
\usepackage{epsfig}
\usepackage{graphicx}
\usepackage{amsmath}
\usepackage{amssymb}
\usepackage{multirow}

\usepackage[numbers,sort]{natbib} 
\usepackage{tabularx} 

\newcommand{\UCOchance}{40.4}
\newcommand{\UCOLAEO}{UCO-LAEO\xspace}
\newcommand{\AVAchance}{17.1}
\newcommand{\AVALAEOscore}{39.1}
\newcommand{\AVALAEOscoreTrAVA}{50.6}
\newcommand{\UCOLAEOscore}{79.5}
\newcommand{\UCOLAEOscoreTrAVA}{77.8}
\newcommand{\TVHIDscoreTrAVA}{90.7}

\let\oldsim\sim 
\renewcommand{\sim}{{\oldsim}}

\usepackage{color}
\usepackage{xcolor}

\renewcommand{\paragraph}[1]{\vspace{2mm} \noindent \textbf{#1}}
\newcommand\paragraphV{\vspace{0.7mm}\paragraph}

\usepackage{enumerate}
\usepackage{enumitem}

\usepackage{}

\usepackage{float}
\usepackage{soul}

\usepackage{pifont}
\usepackage{caption}

 \cvprfinalcopy 

\def\cvprPaperID{1725} 

\ifcvprfinal\fi
\begin{document}

\title{LAEO-Net: revisiting people Looking At Each Other in videos}

\author{
		\hspace*{-12pt}Manuel J. Mar\'in-Jim\'enez\\
		\hspace*{-12pt}University of Cordoba\\
		\hspace*{-12pt}{\tt \small \href{mailto:mjmarin@uco.es}{\textcolor{black}{mjmarin@uco.es}}} 
	\and	
		\hspace*{-10pt}Vicky Kalogeiton\\
		\hspace*{-10pt}University of Oxford\\
		\hspace*{-10pt}{\tt \small \href{mailto:vicky@robots.ox.ac.uk}{\textcolor{black}{vicky@robots.ox.ac.uk}}}
	\and 
		\hspace*{-7pt}Pablo Medina-Su\'arez\\
		\hspace*{-7pt}University of Cordoba\\
		\hspace*{-7pt}{\tt \small  \href{mailto:i42mesup@uco.es}{\textcolor{black}{i42mesup@uco.es}}}
	 \and 
		 \hspace*{-2pt}Andrew Zisserman \hspace*{-15pt}\\
		 \hspace*{-2pt}University of Oxford \hspace*{-15pt}\\
		 \hspace*{-2pt}{\tt \small  \href{mailto:az@robots.ox.ac.uk}{\textcolor{black}{az@robots.ox.ac.uk}}} \hspace*{-15pt}
}

\twocolumn[{
\renewcommand\twocolumn[1][]{#1}
\maketitle
\centering
\includegraphics[width=0.9\linewidth]{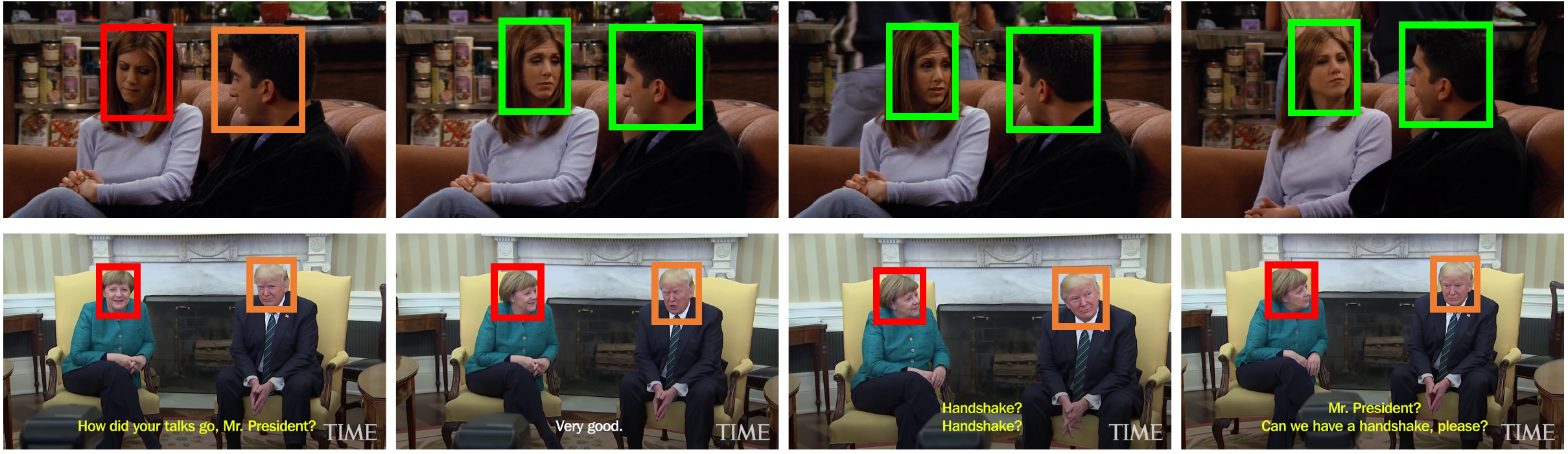}
\captionof{figure}{
\small{\textbf{Intimacy or hostility?} Head pose, along with body pose and facial expressions, is a rich source of information for interpreting human interactions. Being able to automatically understand the non-verbal cues provided by the relative head orientations of people in a scene enables a new level of human-centric video understanding. Green and red/orange heads represent LAEO and non-LAEO cases, respectively. Video source of second row: \url{https://youtu.be/B3eFZMvNS1U} 
   }} 
   \label{fig:teaser}
\vspace{10px}
}]

\begin{abstract}
Capturing the `mutual gaze' of people is essential for understanding and interpreting the social interactions between them. 
To this end, this paper addresses the problem of detecting people \textit{Looking At Each Other (LAEO)} in video sequences. 
For this purpose, we propose LAEO-Net, a new deep CNN for determining LAEO in videos. 
In contrast to previous works, LAEO-Net takes spatio-temporal tracks as input and reasons about the whole track. 
It consists of three branches, one for each character's tracked head and one for their relative position. 
Moreover, we introduce two new LAEO datasets: \UCOLAEO and AVA-LAEO.
A thorough experimental evaluation demonstrates the ability of LAEO-Net to successfully determine if two people are LAEO and the temporal window where it happens. 
Our model achieves state-of-the-art results on the existing TVHID-LAEO video dataset, significantly outperforming previous approaches. 
Finally, we apply LAEO-Net to social network analysis, where we automatically infer the social relationship between pairs of people based on the frequency and duration that they LAEO.
\end{abstract}

\section{Introduction} 
\label{sec:intro}

Eye contact or `mutual gaze' is an important part of the non-verbal
communication between two people~\cite{loeb1972mutual}.  The duration and frequency
of eye contact depends on the nature of the relationship and reflects
the power relationships, the attraction or the antagonism between the
participants~\cite{abele1986gaze}.  
Therefore, in order to understand and interpret the social interactions that are occurring, it is important to capture this signal accurately. 
The importance of detecting people Looking at Each Other (LAEO) has already been recognized in a series of computer vision papers~\cite{marin2011bmvc,marin2013ijcv,palmero2018laeo}  as well as in other papers that study human gaze~\cite{chong2018eccv,recasens2015nips,recasens2017iccv,brau2018eccv}.

LAEO is complementary to other forms of human non-verbal communication such as facial expressions, 
gestures, proxemics (distance), body language and pose, paralanguage (the tone of the voice, prosody), and 
interactions (e.g.\ hugging, handshake). Many of these have been the subject of recent papers \cite{marin2014mva,vondrick2016cvpr,gu2018ava}. 

In this paper, we introduce a new deep convolutional neural network (CNN) for determining LAEO in video
material, coined \textbf{LAEO-Net}. Unlike previous works, our approach answers the question of whether two characters are LAEO over a temporal period by using a spatio-temporal model, whereas previous models have only considered individual frames.  
The problem with frame-wise LAEO is that when characters blink or momentarily move their head, then they are considered non-LAEO, and this can severely affect the accuracy of the LAEO measurement over a time period.
The model we introduce considers head tracks over multiple frames, and determines whether two characters are LAEO for a time period based on the pose of their heads and their relative position. Such an example is in Figure~\ref{fig:teaser}.

We make the following contributions: first, we introduce a spatio-temporal LAEO model that consists of three branches, one for each character's tracked head and one for their relative position, together with a fusion block. 
This is described in Section~\ref{sec:model}. 
To the best of our knowledge, this is the first work that uses tracks as input and reasons about people LAEO in the whole track, instead of using only individual frames.
Second, we introduce two new datasets (Section~\ref{sec:datasets}): 
(i)~{\bf \UCOLAEO}, a new dataset for training and testing LAEO. It consists of  $129$ ($3$-$12$ sec) clips from four popular TV shows; and 
(ii)~{\bf AVA-LAEO}, a new dataset, 
which extends the existing large scale AVA dataset~\cite{gu2018ava} with LAEO annotations for the training and validation sets.  
We evaluate the performance of the spatio-temporal LAEO model on both these new datasets (Section~\ref{sec:expers}).
Third, we show that our model achieves the state of the art on the existing TVHID-LAEO dataset~\cite{marin2013ijcv} by a significant margin ($3\%$). 
Finally, in Section~\ref{sec:friends}, we show that by using LAEO scores we can compute a social network from character interactions in TV material, and we demonstrate this for two episodes of the TV comedy `Friends'.

\begin{figure*}[t]
\centerline{
\includegraphics[width=0.85\linewidth]{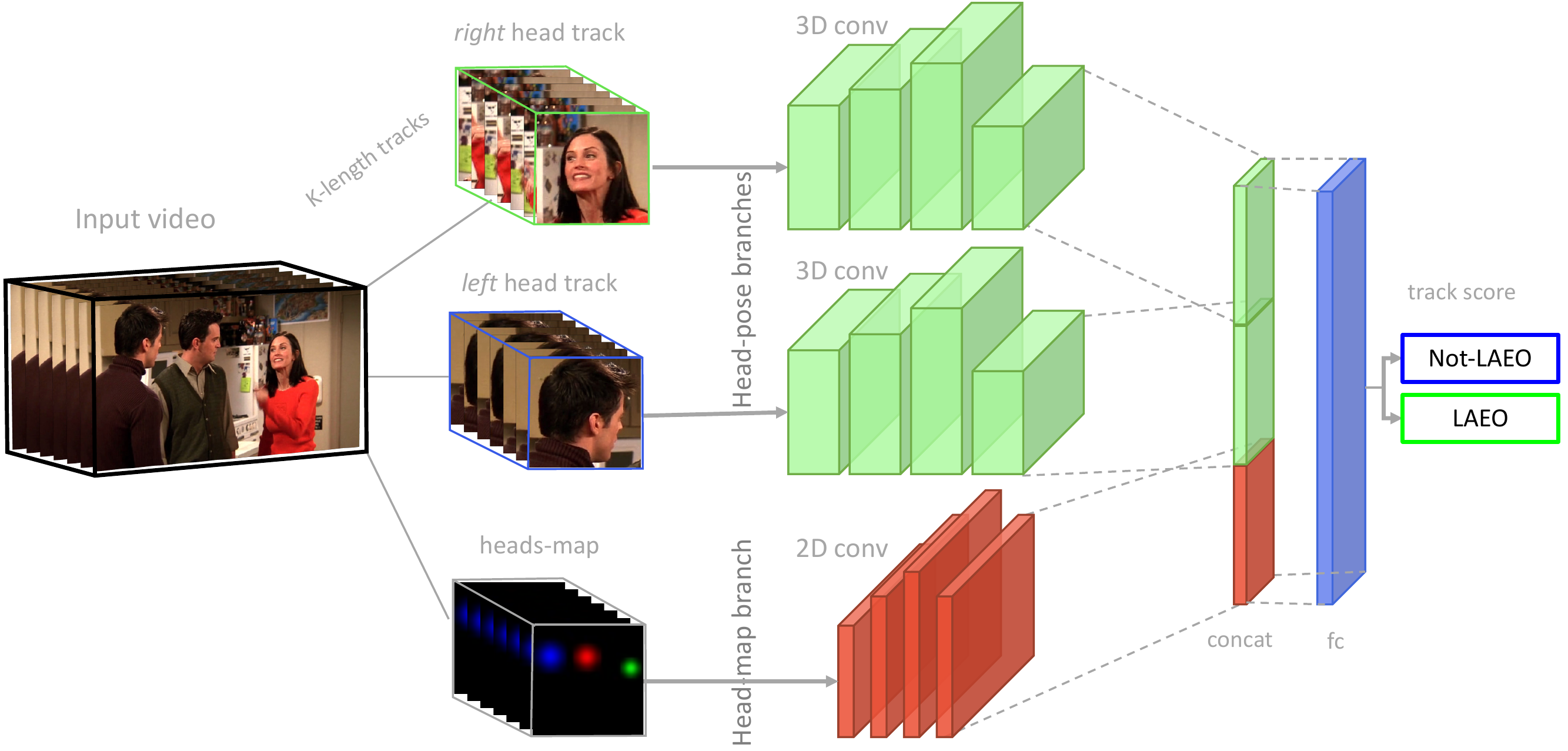}
}
 \caption{\small{\textbf{Our three branch track LAEO-Net}: It consists of the head branches (green), the head-map branch (red) and a fusion block, which concatenates the embeddings from the other branches and scores the track sequence as LAEO or not-LAEO with a fully connected layer (blue) using softmax loss. In our experiments, we use head tracks of length $K=10$.}}
 \vspace{-2mm}
 \label{fig:main}
\end{figure*}
\section{Related work} 
\label{sub:relworks}

Gaze~\cite{recasens2015nips} and head pose~\cite{drouard2017hpose} are powerful tools to deal with the problem of determining the \textit{visual focus of attention} (VFoA) in a scene, \ie what people are looking. 
For instance, \cite{kobayashi2001unique} highlights the importance of the white part of the human eye (\ie white sclera) in recognising gaze direction, enabling the extraordinary ability of humans to communicate between each other just by using gaze signals.

\paragraphV{Visual focus of attention.} 
One classical approach for determining the VFoA is \cite{ba2009vfoa}, where the authors model the dynamics of a meeting group in a probabilistic way, inferring where the participants are looking at. 
More recently, the authors of~\cite{brau2018eccv} discover 3D locations of regions of interest in a video scene by analysing human gaze. 
They propose a probabilistic model that simultaneously infers people's location and gaze as well as the item they are looking at, which might even be outside the image. 
To evaluate their model, they record a total of $8$ videos in controlled scenarios. 

\paragraphV{Gaze direction.}
Instead of videos, the work of~\cite{chong2018eccv} focuses on still images by proposing a neural network that estimates both the gaze direction and the VFoA, even if it is outside the image. A coarse spatial location of the target face in the image is provided to the network in the form of a one-hot vector. 
In contrast, in our model, the relative position of the human heads is provided by creating an RGB image with Gaussian-like circles representing the centre and scale of the heads and a colour-coding for indicating what is the target pair of heads to be analysed (Figure~\ref{fig:headmaps_synpairs}~(a). 
Therefore, our representation offers a better resolution of the scene geometry and incorporates information about head scales, providing a stronger input signal to the network.

Typically, in commercial movies, an action is represented among alternating video shots. Therefore, sometimes the VFoA is not visible in the current frame or shot, but in a different one. This problem is addressed in~\cite{recasens2017iccv} with a deep learning model that reasons about human gaze and 3D geometrical relationships between different views of the same scene. 
The authors of~\cite{masse2018pami} consider scenarios where multiple people are involved in a social interaction. 
Given that the eyes of a person are not always visible (\eg due to camera viewpoint), they estimate people's gaze by modelling the motion of their heads with a Bayesian model, in which both gaze and VFoA are latent variables. 

\paragraphV{People Looking At Each Other. }
A special case of VFoA is when subject-A's VFoA is subject-B, and subject-B's VFoA is subject-A. This case is known as \textit{mutual gaze} or people \textit{looking at each other} (LAEO). 
This situation typically entails a non-physical human interaction; but might precede or continue a physical one, \eg hand-shake before starting a conversation, or after finishing it. 
In the context of \textit{Behaviour Imaging} research area, detecting LAEO events is a key component for understanding higher-level social interactions as, for example, in the study of autism in children~\cite{rehg2011behavior}.

The problem of detecting people LAEO in videos was first introduced in \cite{marin2011bmvc,marin2013ijcv}. After detecting and tracking human heads,  \cite{marin2013ijcv} model and predict yaw and pitch angles of the human heads with a Gaussian Process regression model. 
Based on the estimated angles and the relative position of the two heads a LAEO score was computed per frame, and aggregated to provide a shot-level LAEO score.
Even though we also model the head pose and relative position between two heads, our LAEO-Net differs from these works as it exploits the temporal consistency of neighbouring frames and estimates LAEO events for a track that spans over a temporal window, instead of a single frame.

In a controlled scenario with just two people interacting, \cite{palmero2018laeo} addresses the LAEO problem by using two calibrated cameras placed in front of the two participants, making sure that there is an overlapping visible zone between both cameras. 
It first estimates the eye gaze with a CNN and, then, uses 3D geometry to decide if the ray gaze of either participant intersects the head volume of the other one. 
In contrast, our goal is to address LAEO detection in general scenarios, with a (potentially) unrestricted number of subjects.

\section{LAEO-Net}
\label{sec:model}

\begin{table}[tb]
\centering
\small{
\begin{tabular}{l|l}
\hline
\multicolumn{2}{c}{branch details (`f': filter, `s': stride, $h \times w \times t$) } \\ 
\hline
\multicolumn{1}{c}{head-pose} & \multicolumn{1}{|c}{heads-map}  \\
input: frame crops 64$\times$64$\times$K & input: map 64$\times$64 \\
(four 3D conv layers) & (four 2D conv layers)  \\ 
\hline
f: $16: 5\times5\times3$,  s:$2\times2\times1$ & 
f: $8\times5\times5$,   s:$2\times2$  \\ 
f: $24: 3\times3\times3$,  s:$2\times2\times1$ & 
f: $16\times3\times3$,  s:$2\times2$  \\
f: $32: 3\times3\times3$,  s:$2\times2\times1$ & 
f: $24\times3\times3$,  s:$2\times2$  \\
f: $12: 6\times6\times1$,  s:$1\times1\times1$ &  
f: $16\times3\times3$,  s:$4\times4$  \\ 
\hline
\end{tabular}
}
\caption{\small{\textbf{Architecture specification. }}
}
\label{tab:branches}
\vspace{-4mm}
\end{table}

Given a video clip, we aim to determine if any two humans are \textit{Looking At Each Other} (LAEO). 
To this end, we introduce the LAEO-Net, a three branch \textit{track} network, which takes as input two head tracks and the relative position between the two heads encoded by a head-map, and determines a confidence score on whether the two people are looking at each other or not, and the frames where LAEO occurs. 
The network is applied exhaustively over all pairs of simultaneous head tracks in the video clip.

LAEO-Net consists of three input branches, a fusion block, and a fully-connected layer and is illustrated in Figure~\ref{fig:main}.  Two of the input streams determine the pose of the heads (green branches) and the third represents their relative position and scale (red branch). 
The fusion block combines the embeddings from the three branches and passes them through a fully-connected layer that predicts the LAEO classification (blue layer). 
The network uses spatio-temporal 3D convolutions and can be applied to the head tracks in the video clip. 
We next describe the components in detail and report their specifications in Table~\ref{tab:branches}.

\paragraph{Head-pose branch. }
It consists of two branches, one per person. 
The input of each branch is a tensor of $K$ RGB frame crops of size $64 \times 64$ pixels, containing a sequence of heads of the same person.
Each branch encodes the head frame crop, taking into account the head pose.
The architecture is inspired by the one proposed in \cite{gygli2017shots} for shot boundary detection, with four 3D conv layers followed by a dropout and a flatten ones (green branches in Figure~\ref{fig:main}). 
The output of the flatten layer is L2-normalized before using it for further processing.
Note that the head sequence of each person of the target pair will be encoded by this branch, obtaining two embedding vectors as a result.

\begin{table*}[ht]
\begin{minipage}[b]{0.7\textwidth}\centering
    \small{
    \begin{tabular}{ll|r|r|r|r|r}
    \hline
    \multicolumn{7}{c}{Overview of LAEO datasets} \\
    \hline
    \multicolumn{2}{l|}{\multirow{2}{*}{statistics}}& \multicolumn{2}{c|}{\UCOLAEO (\textbf{new})} & \multicolumn{2}{c|}{AVA-LAEO (\textbf{new})} &
    \multicolumn{1}{c}{TVHID-LAEO~\cite{marin2013ijcv}} \\
    \cline{3-7}
    & & \multicolumn{1}{c|}{train+val} & \multicolumn{1}{c|}{test} & \multicolumn{1}{c|}{train} & \multicolumn{1}{c|}{val} & \multicolumn{1}{c}{test} \\
    \hline
    \hline
    \multicolumn{2}{l|}{\#frames} & \multicolumn{2}{c|}{$>18$k} & \multicolumn{2}{c|}{$>1,4$M (estim.)} & \multicolumn{1}{c}{$>29$k} \\ 
    \multicolumn{2}{l|}{\#programs} & \multicolumn{2}{c|}{4 (tv-shows)} & \multicolumn{2}{c|}{298 (movies)} & \multicolumn{1}{c}{20 (tv-shows)}\\
    \hline
    \multirow{2}{*}{shots} & \multicolumn{1}{|l|}{\#annotations} & 106+8 & 15  &  40166 & 10631  & 443  \\ 
    & \multicolumn{1}{|l|}{\#LAEO} & 77+8 & 15 & 18928 & 5678  & 331  \\ 
    \hline
    \multirow{2}{*}{pairs} & \multicolumn{1}{|l|}{\#annotations} & 27358+5142 & 3858 & 137976 & 34354  & -- \\ 
    & \multicolumn{1}{|l|}{\#LAEO} & 7554+1226 & 1558 & 19318 & 5882  & -- \\ 
    \hline
    \multicolumn{2}{l|}{sets (pairs)} & 32500 & 3858 & 137976 & 34354 & 443  (shots)\\
     \hline
    \end{tabular}
    }
    \captionof{table}{\small{\textbf{Summary of LAEO datasets.}   \textit{\#programs}:  different TV shows; \textit{{\#}shot-annotations}: annotated shots; \textit{{\#}shot-LAEOs}: shots containing at least one LAEO pair; \textit{{\#}pair-annotations}: annotated human bounding box pairs; \textit{{\#}pair-LAEOs}: human bounding box pairs that are LAEO; \textit{sets:} {\#}train/val/test LAEO pairs (or shots) used.}}
    \label{tab:dbstats}
\end{minipage}
\hfill
\begin{minipage}[b]{0.29\textwidth}
    \centering
    \includegraphics[width=0.87\textwidth]{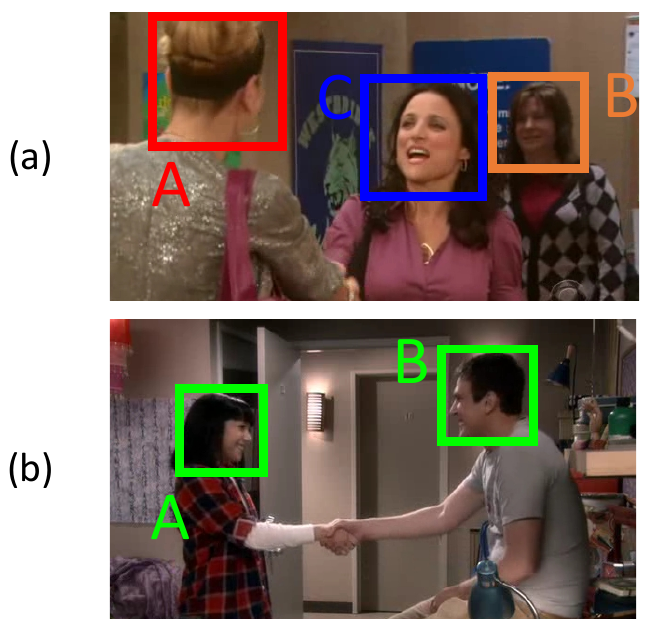}
    \vspace{4mm}
    \captionsetup{skip=0cm, width=0.8\textwidth}
    \captionof{figure}
    {
    		\small{
    			(a) AB are not LAEO as C is occluding. (b) AB are LAEO.}
    }
    \label{fig:whyhmap}
\end{minipage}
\end{table*}


\paragraph{Head-map branch. }
This branch embeds the relative position between two head tracks over time using a head-map. 
In particular, we define a $64 \times 64$ map to depict as 2D Gaussians all the heads detected in the central frame of the $K$-frames track (Figure~\ref{fig:headmaps_synpairs}~(a)). 
In addition to the two head tracks, this branch encodes information for other people in the scene. 
Depending on its size and scale, a third person could cut the \textit{gaze ray} between the two side people (Figure~\ref{fig:whyhmap}). 
Including this information helps the LAEO-Net to distinguish such cases. 
This branch consists of a four 2D conv layers  (Table~\ref{tab:branches}). 
To obtain the embedding of the head-map we flatten the output of the last conv layer and apply L2-normalization. 

\paragraph{Fusion block.} 
The embedding vectors obtained as the output of the different branches of the network are concatenated and further processed by one fully-connected layer with an alternating dropout layer (blue layer in Figure~\ref{fig:main}). 
Then, a Softmax layer consisting of two hidden units (\ie representing not-LAEO and LAEO classes) follows.

\paragraph{LAEO loss function. }
For training the LAEO predictor, we use the standard binary cross entropy loss: 
\begin{equation}
\label{eq:laeoLoss}
\small{{\mathcal{L}_{\textrm{LAEO}} = - \left( c \cdot \log (\hat{p}_{c}) + (1-c) \cdot \log(1-\hat{p}_{c}) \right),}}
\end{equation}
where $c$ is the ground-truth class (\ie $0$ for not-LAEO, $1$ for LAEO) and $\hat{p}_{c}$ is the predicted probability of the pair being class $c$.

\section{Datasets} 
\label{sec:datasets}

In this section, we describe the LAEO datasets. 
First, we introduce two new datasets: \UCOLAEO and AVA-LAEO, and 
then, two other datasets: AFLW~\cite{koestinger11aflw}, and TVHID~\cite{Patron2010hi5}. 
AFLW is used for pre-training the head-pose branch and for generating synthetic data, while TVHID is used only for testing. 
The newly introduced \UCOLAEO and AVA-LAEO datasets are used both for training and testing the LAEO-Net. 
Table~\ref{tab:dbstats} shows an overview of the LAEO datasets.

The new datasets with their annotations and the code for evaluation are available online at:  \url{http://www.robots.ox.ac.uk/~vgg/research/laeonet/}.

\subsection{The \UCOLAEO dataset }
\label{sub:dat_laeo}
We use four popular TV shows: `Game of thrones', `Mr Robot', `Smallville' and `The walking dead'. 
From these shows, we collect $129$ ($3$-$12$ seconds long) shots and first annotate all the heads in each frame with bounding boxes, and then annotate each head pair as LAEO or not-LAEO. 
Some examples are shown in Figure~\ref{fig:datasets}~(top).

\paragraph{Annotation setup. } 
We annotate all frames both at the frame level, \ie, \textit{does this frame contain any pair of people LAEO?}; and  
at the head level, \ie we annotate all heads in a frame with a bounding-box and all the possible LAEO pairs. 
The visually ambiguous cases are assigned as `ambiguous' and we exclude them from our experiments. 
Some examples are shown in Figure~\ref{fig:datasets}~(top). 
We split the $100$ LAEO shots into $77$ train, $8$ validation and $15$ test, respectively. This results in $\sim7.5$k training, $\sim1.2$k val and $\sim1.5$k test LAEO pairs (Table~\ref{tab:dbstats}).

\subsection{AVA-LAEO dataset }
\label{sub:dat_ava}
AVA-LAEO consists of movies coming from the training and validation sets of the `Atomic Visual Actions' dataset (AVA v2.2)~\cite{gu2018ava} dataset. 
The AVA frames are annotated (every one second) with bounding-boxes for $80$ atomic actions, without containing LAEO annotations; therefore, we enhance the labels of the existing (person) bounding-boxes in a subset of the train and val sets with LAEO annotations. 

\paragraph{Annotation setup. }
From the train and val sets of AVA, we select the frames with more than one person annotated as \textit{`watch (a person)'}, resulting in a total of $40,166$ and $10,631$ frames, respectively. 
We only consider the cases, where both the watcher and the watched person are visible (since the watched person may not be visible in the frame).

\begin{figure*}[t!]
\begin{center}
\includegraphics[width=0.96\textwidth]{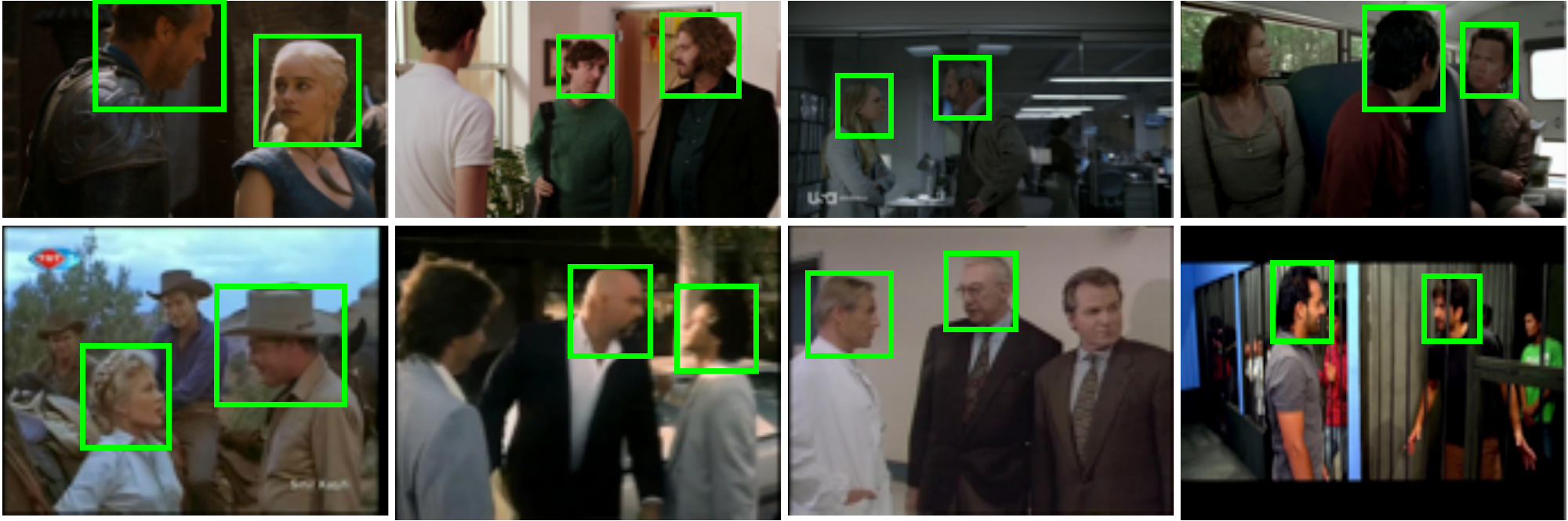}
\end{center}
   \caption{\small{\textbf{(top) \UCOLAEO and (bottom) AVA-LAEO datasets.} 
      Example of frames and LAEO head pair annotations included in our new datasets. Different scenarios, people clothing, background clutter and diverse video resolutions, among other factors, make them challenging.
   }}
\label{fig:datasets}
\vspace{-3mm}
\end{figure*}

For annotating, we follow the same process as in UCO-LAEO, \ie we annotate each pair of human bounding boxes at the frame level as LAEO, not-LAEO, or ambiguous. 
This results in $\sim19$k LAEO and $\sim118$k not-LAEO pairs for the training set and $\sim5.8$k LAEO and $\sim28$k not-LAEO pairs for the val set (Table~\ref{tab:dbstats}). 
We refer to this subset as AVA-LAEO. 
Figure~\ref{fig:datasets}~(bottom) shows some LAEO pair examples. 

\subsection{Additional datasets}

\paragraph{AFLW dataset. } 
\label{subsec:aflw}
We use the `Annotated Facial Landmarks in the Wild' dataset~\cite{koestinger11aflw} to 
(a) pre-train the head pose branch (first stage, Section~\ref{sub:pretrain}), and 
(b) generate synthetic data for training (second stage, Section~\ref{sub:finetune}). 
It contains about $25$k annotated faces in images obtained from FlickR, where each face is annotated with a set of facial landmarks. 
From those landmarks, the head pose (\ie yaw, pitch and roll angles) is estimated. 
To create a sequence of head-crops, we replicate the input image $K$ times. We keep the two middle replicas unchanged and randomly perturbing the others, \ie small shift, zooming and brightness change.

\paragraph{TVHID-LAEO. } 
\label{subsec:tvhid}
The TVHID dataset~\cite{Patron2010hi5} was originally designed for the task of human interaction recognition in videos. 
It contains $300$ video clips with five classes: hand-shake, high-five, hug, kiss and negative. 
We use the LAEO annotations at the shot level from \cite{marin2013ijcv}, which result in $443$ shots with $331$ LAEO and $112$ not-LAEO pairs (Table~\ref{tab:dbstats}). 

\section{Training LAEO-Net}
\label{sec:training}

LAEO-Net requires head tracks as input. 
Here, we first describe the head track generation (Section~\ref{sub:d_t}). 
Then, we describe our training procedure, which involves training in two stages. 
In the first stage (Section~\ref{sub:pretrain}), we pre-train the head-pose branches and freeze their weights. In the second stage, we train the whole LAEO-Net (\ie head-map branch and upper layers) from scratch (Section~\ref{sub:finetune}).

\subsection{Head detection and tracking} 
\label{sub:d_t}
\paragraph{Head detection.}
Our method requires head detections. 
In the literature, there are several models for face detection (\cite{Viola2001} \cite{zhu2012cvpr}); head detection, however, is a more challenging task, as it comprises detecting the whole head, including the face (if visible) but also the back of a head (\eg \cite{marin2013ijcv}).
We train a head detector using the Single Shot Multibox Detector (SSD) detector~\cite{liu2016ssd}\footnote{Detector: \url{https://github.com/AVAuco/ssd_people}} from scratch. 
We train the model with a learning rate of $10^{-4}$ (first $50$ epochs), and decrease it with a factor of $0.1$ for the rest of the training. 
For speedup and better performance we use batch normalization and for robustness we use the data augmentation process from~\cite{liu2016ssd}. 
We train the head detector with the `Hollywood heads' dataset~\cite{vgghollywoodheadsdb}. It consists of head annotations for $1120$ frames, split into $720$ train and $200$ val and test frames. 
We first train our detector with the training set and after validating the model, we train on the whole dataset as a refining stage.

\paragraph{Head tracking. }
Once we obtain head detections, we group them into tracks along time. 
For constructing tracks, we use the online linking algorithm of~\cite{singh17iccv}, as it is robust to missed detections and can generate tracks spanning different temporal extents of the video. 
Out of all head detections, we keep only the $N\!=\!10$ highest scored ones for each frame. 
We extend a track $T$ from frame $f$ to frame $f+1$ with the detection $h_{f+1}$ that has the maximum score if it is not picked by another track and $\text{ov}\left(h_{f}^{T},h_{f+1}\right)\!\geqslant\!\tau$, where $\text{ov}$ is the overlap. 
If no such detection exists for $M$ consecutive frames, the track stops; otherwise, we interpolate the head detections. 
The score of a track is defined as the average score of its detections. 
At a given frame, new tracks start from not-picked head detections. 
To avoid shifting effects in tracks, we track both forwards and backwards in time.

\subsection{Pre-training the head-pose branches} 
\label{sub:pretrain}
In general, humans can infer \textit{where} a person is looking just based on the head pose, without even seeing the eyes of the person. 
This shows that most information is encoded in the head orientation. 
There are several works that model the head orientation \cite{ruiz2018hpose} or the eye gaze~\cite{recasens2015nips}.  
Note that using the actual eye gazing is not always an option, even with multiple-frames as input, as there is no guarantee that the eyes are fully visible, \ie due to image resolution, or self occlusions. Therefore, in this work, we model gaze just based on head orientation. 

We model head orientation with three angles (in order of decreasing information): (a)~the yaw angle, \ie looking right, left, (b)~the pitch angle, \ie looking up, down, and (c)~the roll angle, \ie in-plane rotation. 
We use this modelling to pre-train the head-pose branches. 
We use the weights learnt from this pre-training without tuning this branch further (freezing the weights).

\paragraph{Loss function of head-pose pre-training. }
Let $(\alpha, \beta, \gamma)$ be the yaw, pitch and roll angles of a head, respectively. 
We define one loss for estimating each pose angle: $\mathcal{L}_{\alpha}$, $\mathcal{L}_{\beta}$, $\mathcal{L}_{\gamma}$ and model them with the $L1$-smooth loss~\cite{ren2015fastrcnn}. 

Given that the yaw angle is the dominant one, in addition to these losses, we include a term that penalizes an incorrect estimation of the sign of the yaw angle, \ie, failing to decide if the person is looking left or right ($\mathcal{L}_{s}$). It is defined as: 
\begin{equation}
\mathcal{L}_{s} = \max(0, - \mathrm{sign}(\alpha) \cdot \mathrm{sign}(\hat{\alpha}) ) , 
\end{equation}
where $\mathrm{sign}(\alpha)$ is the sign function (\ie returns $+1$ for positive inputs, $-1$ for negative inputs, and $0$ if the input is $0$) applied to the yaw angle; and, $\hat{\alpha}$ is the ground-truth angle.

Therefore, the loss function $\mathcal{L}_h$ for training the head-pose branch for LAEO purposes is given by:
\begin{equation} \label{eq:headloss}
    \mathcal{L}_h = w_{\alpha} \cdot \mathcal{L}_{\alpha} + w_{\beta} \cdot \mathcal{L}_{\beta} + w_{\gamma} \cdot \mathcal{L}_{\gamma} + w_s \cdot \mathcal{L}_s, 
\end{equation}
where $w_x$ are positive weights chosen through cross-validation during training.
In our experiments, we use: $w_{\alpha} = 0.6$, $w_{\beta} = 0.3$, $w_{\gamma}=0.1$, $w_s=0.1$, as $w_{\alpha}$ is the dominant one determining the head orientation. 
Note that the weights do not necessarily add to $1$.

\subsection{Training the LAEO-Net} 
\label{sub:finetune}

We train the LAEO-Net with both real and synthetic data. 
We also use \textbf{data augmentation} techniques, such as image perturbations, translations, brightness changes, zoom changes, \etc. 
For the first $N=2$ epochs, we train the LAEO-Net only with synthetic data, and then at each \textbf{training step} we alternate between real and synthetic data. 
To improve the performance of the model, we also use hard negative mining. 
We deploy the \textbf{curriculum learning} strategy of \cite{Nagrani18c}, which facilitates the learning by modulating the difficulty of the hard negatives incorporated into the training phase. 
In our experiments, the value of the hyper-parameter $\tau$ is increased after $2$ epochs, allowing more difficult examples as its value increases. 
The importance of our training scheme is evaluated in Section~\ref{sub:synthetic}.

\begin{figure}[t!]
\centerline{%
\begin{tabular}{c@{}c@{}}
\includegraphics[width=0.52\linewidth]{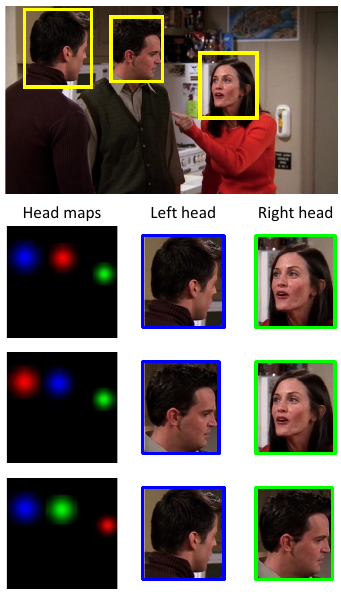}& 
\includegraphics[width=0.46\linewidth]{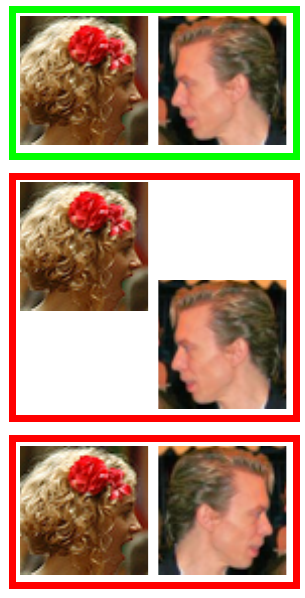} \\
\small{(a)} & \small{(b)} \\
\end{tabular}}
\caption{\small{\textbf{(a)~Head-maps and (b)~augmentation of LAEO training samples.} 
(a)~For each frame, we analyse all head pairs, using a color coding: \textit{blue} for the left, \textit{green} for the right and \textit{red} for the remaining heads, such as middle (\ie not currently considered for evaluation).  
(b)~We generate additional synthetic LAEO negative training data (red boxes) from each positive pair (green box), based on the orientation or the relative position of the heads. 
}}
\label{fig:headmaps_synpairs}
\vspace{-3mm}
\end{figure} 

\paragraph{Synthetic data. }
For generating additional synthetic data  we use images with head-pose information. 
To generate positive samples, we select pairs of head images whose angles are compatible with LAEO and, at the same time, they generate consistent geometrical information. 
To generate negative samples, we 
(i)~change the geometry of the pair, \ie making LAEO not possible any more, for instance, by mirroring just one of the heads of the pair or 
(ii)~select pairs of head images whose pose are incompatible with LAEO, \eg both looking in the same direction. 
Figure~\ref{fig:headmaps_synpairs}~(b) shows some artificially generated examples.

\section{Experimental results}
\label{sec:expers}
%
In this section, we experimentally evaluate the effectiveness of LAEO-Net for determining people LAEO. 
Note that the model is trained either on the \UCOLAEO or on the AVA-LAEO dataset. 
We, first, report the evaluation protocol (Section~\ref{sub:metrics}), and we assess the importance of synthetic data during fine-tuning (Section~\ref{sub:synthetic}). 
Then, we examine the impact of each component of the model (Section~\ref{sub:ablation}), and in Section~\ref{sub:res_newdatasets} we examine the performance of LAEO-Net on the two new test datasets, \UCOLAEO and AVA-LEO.  
Finally, we compare our LAEO-Net to the state-of-the-art methods on the TVHID-LAEO dataset (Section~\ref{sub:restvhid}). 

\paragraph{Implementation details.}
LAEO-Net is implemented with Keras~\cite{chollet2015keras} using TensorFlow as backend. 
For training the LAEO-Net we use the Adam optimizer with mini-batches of 9 samples: 4 positive, 4 negative and 1 hard negative. 
The learning rate starts at $10^{-4}$ and is reduced by a factor of $0.2$ when the AP on the validation set does not improve for 5 consecutive iterations. The minimum possible learning rate is set to $10^{-8}$. 
Dropout is set to $0.5$. 
Before applying the trained model to the test sets, the samples of the validation set of \UCOLAEO are added to the training set for few additional epochs using the latest learning rate.

\subsection{Evaluation protocols and scoring methodology}
\label{sub:metrics}

\paragraph{LAEO-classification AP} 
is the metric we use to evaluate the LAEO predictions. 
Similar to object detection, a detection is correct if its intersection-over-union (IoU) overlap with the ground-truth box is $>0.5$ \cite{voc}. 
A detected pair is correct if both heads are correctly localized and its label (LAEO, not-LAEO) is correct. 
The performance is Average Precision (AP) computed as the area under the Precision-Recall (PR) curve. 
Depending on the available ground-truth annotations, we measure AP at frame level, considering each pair as an independent sample, or at shot-level, if more detailed annotations are not available. 
Frame level is used for \textit{\UCOLAEO} and \textit{AVA-LAEO} and, following previous work~\cite{marin2013ijcv,masse2018pami}, shot level for \textit{TVHID}.

\paragraph{Scoring methodology. }
Given that the level of (ground truth) annotation differs between the three datasets, we describe how we use the LAEO-Net outputs to obtain the final scores (either at the shot or at the frame level). 
To this end, we test the LAEO-Net on pairs of head-tracks (of length $K=10$) and obtain one LAEO score for each track-pair. We assign the LAEO score to the head-pair in the middle frame. 
We describe below the scoring process for each dataset.

\noindent i. \textit{\UCOLAEO:} 
Since the bounding boxes for the heads are available for each frame, the LAEO-Net is applied directly to these head tracks (no detections are used). 
To account for the $K/2$ frames at the beginning (resp.\ end) of a track, we propagate the score from the middle frame.

\noindent ii. \textit{AVA-LAEO:}
We run the head tracker and apply the LAEO-Net on these tracks. 
Given that AVA-LAEO contains pair annotations for \textit{human} bounding-boxes (instead of heads), we compare each detected head pair against the ground-truth human pairs using intersection over head area (instead of IoU).

\noindent iii. \textit{TVHID:}
We run the head tracker and apply the LAEO-Net on these tracks. We compute a LAEO score as the maximum of the smoothed scores in a shot; 
the smoothed score of each pair is the average of a moving temporal window (of length five) along the track.

\subsection{Importance of synthetic data}
\label{sub:synthetic}
Using synthetic data (\ie change of relative position of heads to create LAEO and not-LAEO pairs, see Figure~\ref{fig:headmaps_synpairs}~(b) allows for more training samples for free, thus making the model more generalizable while reducing the probability of overfitting (Section~\ref{sub:finetune}). 
Training and testing LAEO-Net on \UCOLAEO results in an AP=$\UCOLAEOscore\%$, while training: 
\noindent (i)  without synthetic data and without hard negative mining results in AP=$64.8\%$, \ie $14.7\%$ decrease,  
\noindent (ii) with synthetic data but without hard negative mining results in AP=$70.2\%$, \ie $9.3\%$ decrease, 
\noindent (iii) without synthetic data but with hard negative mining results in AP=$71.2\%$, \ie $8.2\%$ decrease,  
\noindent (iv) with only synthetic data (no real data and no hard negative mining) results in AP=$76.9\%$, \ie $2.6\%$ decrease.
These results reveal the significance of using synthetic data in the training process. 

\begin{table}
\centering
\footnotesize{{
\begin{tabular}{c||c|c|c}
\hline
head pose +  & \multicolumn{3}{c}{LAEO \%AP on \UCOLAEO}  \\
\hline
head-map & K=1 (2D) & K=5 & \textbf{K=10}  \\

\hline
\hline
-          &       64.9 &       58.7 & 73.5 \\ 
\checkmark &      72.7  &  73.9 & \textbf{\UCOLAEOscore} \\ 
\hline
\end{tabular}}}
\vspace{-1mm}
\caption{\small{\textbf{Ablation study of LAEO-Net}. We report \%AP of LAEO-Net when trained and tested on \UCOLAEO for various temporal windows K=1,5,10.
}}
\label{tab:ablation}
\vspace{-4mm}
\end{table}

\subsection{Ablation study} 
\label{sub:ablation}
LAEO-Net consists of the head-pose and the head-maps branches. 
Here, we study the impact of some architecture choices, in particular, the head-maps branch and the length of the temporal window $K$ (Table~\ref{tab:ablation}).

We evaluate the LAEO-Net without and with the head-map branch. 
We observe that adding the heap-map branch improves the performance of our architecture (from $73.5\%$ to $\UCOLAEOscore\%$ for $K=10$), as it enables learning the spatial relationship between the two heads.  
Moreover, to assess the importance of the temporal window using $K$ frames compared to using a single frame, we vary $K$ and train and evaluate the LAEO-Net with $K=1,5,10$. 
Table.~\ref{tab:ablation} shows that there is an improvement in AP performance (of $1.2\%$) when K increases from only 1 to 5 frames, and a significant improvement (of $6.8\%$) when K increases from only 1 to 10 frames.
In this work, we use $K=10$ frames. 
Training the LAEO-Net without freezing the weights of the head-pose branch results in AP=$75.1\%$ vs AP=\UCOLAEOscore\%, demonstrating that freezing weights results in performance improvements.

\paragraph{Comparison with an alternative architecture. }
\label{sub:alternatives}
The core branches of LAEO-Net are the head-pose branches. 
In addition, we use the head-map branch to describe the relative position between two heads in the tracks over time. 
Here, we consider one alternative to the head-map, the \textit{geometrical information branch}, where the relative position of two heads over time is encoded based on their geometry.

The \textit{geometrical information branch} embeds the relative position between two head tracks over time (relative to a $(1,1)$ normalized reference system), as well as the relative scale of the head tracks. The input is a tuple $(dx, dy, s_{r})$, where $dx$ and $dy$ are the $x$ and $y$ components of the vector that goes from the left head $L$ to the right one $R$, and $s_{r} = s_{L}/s_{R}$, is the ratio between the scale of the left and right heads. 
The encoding network consists of two fully-connected layers with 64 and 16 hidden units, respectively. 
Therefore, it outputs a vector of 16 dimensions encoding the the geometrical relation between the two target heads. 

When we replace the head-map branch with the geometry branch, 
for \UCOLAEO and $K=10$, the classification AP of the LAEO-Net with the \textit{geometry branch} is 
is between $3-5\%$ less than the AP with the head-map branch. 
This is expected; even though both branches encode the same information (\ie relative position of the two heads), the head-maps branch provides a richer representation of the scene and, therefore, results in better AP. 
Note that using both the head-map \textit{and} the geometry branches (in addition to the head-pose branches) does not lead to any further performance improvement, due to the fact that the combination of these two branches just increases the number of network parameters without providing any additional information. 
Therefore, we conclude that the proposed LAEO-Net is the most effective architecture in terms of AP performance.

\begin{table}
\centering
\small
\resizebox{\linewidth}{!}{
\begin{tabular}{l||r|r||r|r||r|r|r}
\hline
\multicolumn{8}{c}{LAEO \% AP}  \\ \hline 
train on & UCO & AVA & UCO & AVA & UCO & AVA & TVHID \\
\hline 
test on & \multicolumn{2}{c||}{\UCOLAEO} & \multicolumn{2}{c||}{AVA-LAEO} & \multicolumn{3}{c}{TVHID} \\ 
\hline 
baseline (chance level) & \multicolumn{2}{c||}{\UCOchance} & \multicolumn{2}{c||}{\AVAchance} & \multicolumn{3}{c}{--} \\ 
\cite{marin2013ijcv} (Fully auto+HB)  & -- & -- & -- & -- & -- & -- & 87.6 \\ 
\cite{masse2018pami} (Fine head orientation) & -- & -- & -- & -- &  -- & -- & 89.0 \\ 
LAEO-Net & \textbf{\UCOLAEOscore}  & \UCOLAEOscoreTrAVA & \AVALAEOscore & \textbf{\AVALAEOscoreTrAVA} & \textbf{91.8} & \TVHIDscoreTrAVA & -- \\
\hline
\end{tabular}}
\vspace{-1.5mm}
\caption{\small{\textbf{LAEO results on \UCOLAEO, AVA-LAEO and TVHID.} We report \%AP at the pair@frame level for TV-LAEO and AVA-LAEO and, similar to other works, at the shot level for TVHID. 
}}
\vspace{-4.5mm}
\label{tab:results}
\end{table}

\subsection{Results on \UCOLAEO and AVA-LAEO}
\label{sub:res_newdatasets}
We evaluate the LAEO-Net on the \UCOLAEO and AVA-LAEO datasets and report the results in Table~\ref{tab:results}. 
When training and testing on \UCOLAEO, the performance is $\UCOLAEOscore\%$, demonstrating the effectiveness of our model. 
When training and testing on AVA-LAEO, the performance is $\AVALAEOscoreTrAVA\%$. 
These results reveal that there exists a significant gap in the performance between \UCOLAEO and AVA-LAEO. 
This is due to the different nature of AVA-LAEO compared to other datasets: 
(1) head annotations are not provided (just human bounding-boxes every 1 second); 
(2) it contains challenging visual concepts, such as (a) low resolution movies, (b) many people in a scene, (c) blurry, small heads, and (d) particular clothing styles, \eg several people wearing hats (western, Egyptian's, turbans, \etc). 
Despite these difficulties, the LAEO-Net achieves AP=$\AVALAEOscoreTrAVA\%$ on AVA-LAEO.

To examine the generalization of LAEO-Net to other datasets, we also 
report results when training and testing with different datasets, \ie, AP=$\UCOLAEOscoreTrAVA\%$ for \UCOLAEO and AP=$\AVALAEOscore\%$ for AVA-LAEO. 
These results show that the domain shift~\cite{torralba11cvpr} definitely affects the performance, 
but at the same time our models are able to generalize to other 
unseen datasets. 
To assess the difficulty of these
datasets and the effectiveness of LAEO-Net, we also compare it to the chance level classification, 
LAEO-Net outperforms chance level by a large margin: $+40\%$ for \UCOLAEO and $+33\%$ for AVA-LAEO.  

\begin{figure}[t!]
\begin{center}
\includegraphics[width=0.93\linewidth]{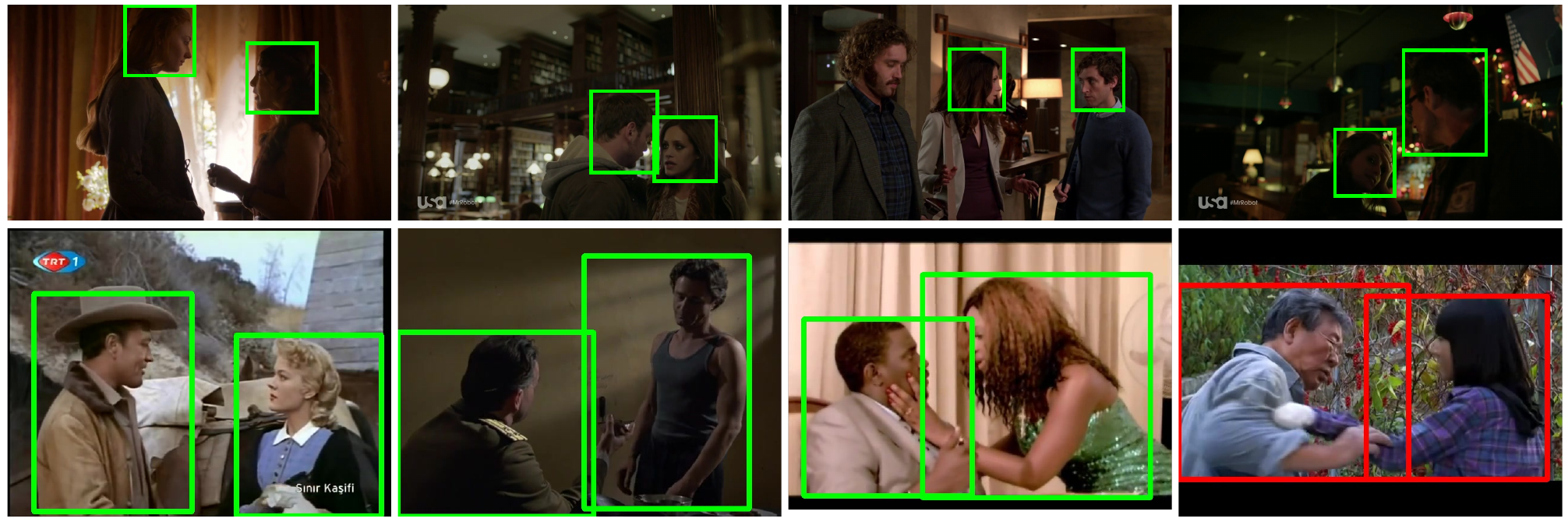}
\end{center}
\vspace{-5mm}
   \caption{\small{\textbf{LAEO-Net results on (top) \UCOLAEO and (bottom) AVA-LAEO.} 
    For different scenarios, backgrounds, head poses \etc, in most cases the LAEO-Net successfully determines if two people are LAEO (green boxes); in some other cases (red boxes), only by tracking the eye movement, we could determine if the people are LAEO. 
    }}
\label{fig:res_datasets}
\vspace{-3mm}
\end{figure}

When applying LAEO-Net on \UCOLAEO and AVA-LAEO we obtain the results of Figure~\ref{fig:res_datasets}.  
For both datasets, we display some of the highest ranked pairs of people that are LAEO. 
We observe that the LAEO-Net leverages the head orientations and their temporal consistency and accurately determines the frames where people are LAEO. 

When applying LAEO-Net on \UCOLAEO and AVA-LAEO we obtain the results of Figure~\ref{fig:res_datasets}.  
For both datasets, we display some of the highest ranked pairs of people that are LAEO. 
We observe that the LAEO-Net leverages the head orientations and their temporal consistency and accurately determines the frames where people are LAEO. 

We hope that the LAEO-Net with these two datasets will provide solid baselines and help future research on this area. 

\subsection{Results on TVHID-LAEO}
\label{sub:restvhid}
We compare LAEO-Net to the state of the art on TVHID~\cite{Patron2010hi5}, \ie the only video dataset with LAEO annotations (Section~\ref{subsec:tvhid}). 
As in \cite{marin2013ijcv}, we use average AP over the two test sets of TVHID (Table~\ref{tab:results}). 
On this dataset, LAEO-Net trained on \UCOLAEO and AVA-LAEO achieves AP$=91.8\%$ and AP=$\TVHIDscoreTrAVA\%$, respectively. 
Both these results outperform all other methods by a large margin ($2-3\%$).

\begin{figure}[t]
\begin{center}
\includegraphics[width=0.95\linewidth]{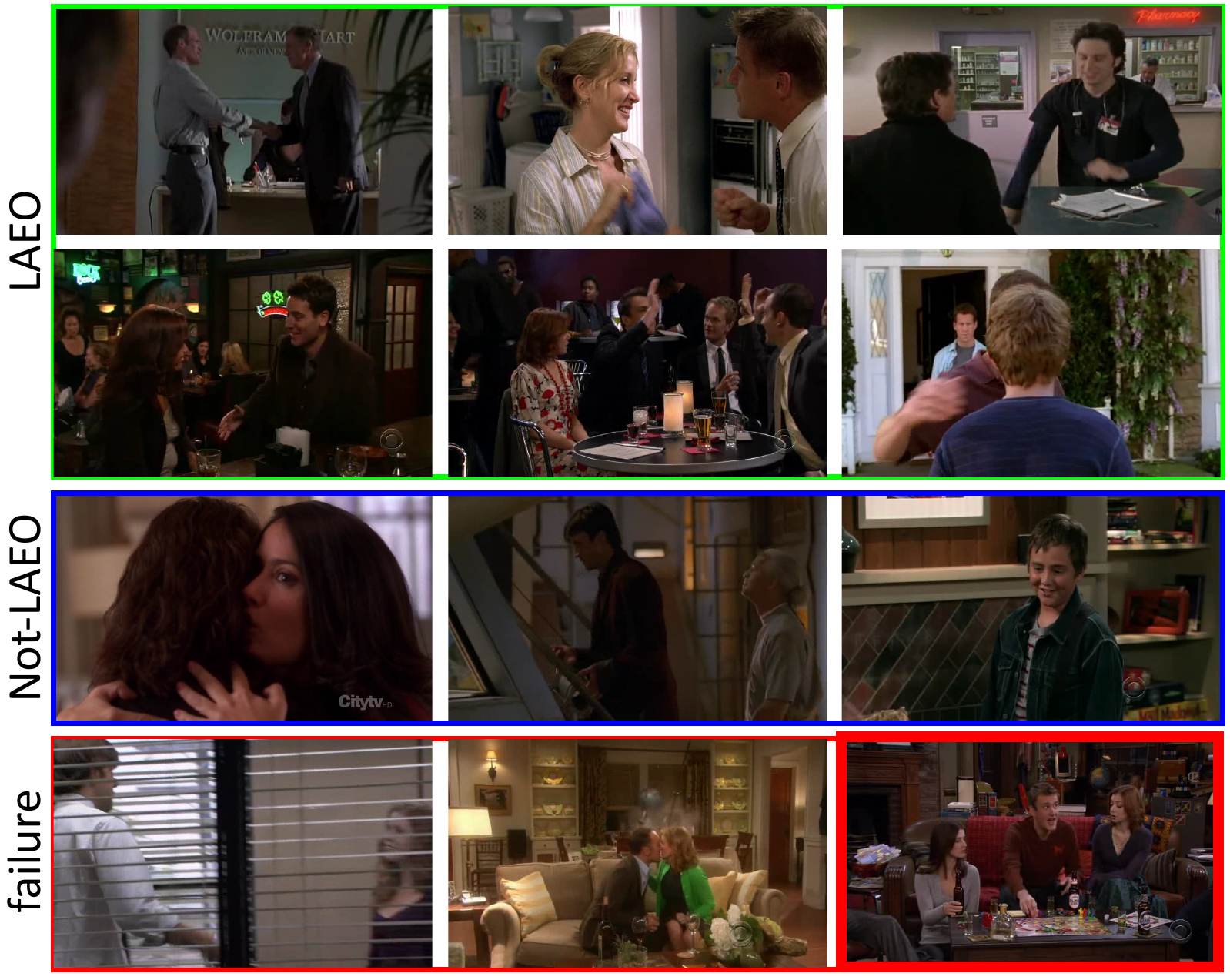}
\end{center}
\vspace{-4mm}
   \caption{\small{\textbf{LAEO-Net results on TVHID.}  
   The top three rows depict correct LAEO results when the ground truth is LAEO (green) and not-LAEO (blue).  
   LAEO-Net successfully detects people LAEO in several situations (different illuminations, scales, cluttered background). 
    Most failure cases are missing people LAEO in ambiguous scenes (first two red frames). 
   In the last red frame though we see a difficult failure case, where LAEO-Net predicts a not-LAEO sequence as LAEO. Note that the head orientation and posture points to LAEO; the character, however, rolls his eyes! 
   }}
\label{fig:res_tvhid}
\vspace{-4mm}
\end{figure}

We apply the LAEO-Net on TVHID and obtain the results shown in Figure~\ref{fig:res_tvhid}. 
Our model successfully detects people LAEO in several situations and scenarios, such as different illuminations, scales, cluttered background. 
By examining the remaining~$8\%$ error, we note that in most cases, the ground truth label is ambiguous (first two red frames in Figure~\ref{fig:res_tvhid}). 
In some cases though, the head pose and relative position are not sufficient cues to model LAEO cases, because the LAEO event can be determined only by examining the eye gaze (last red frame in Figure~\ref{fig:res_tvhid}). 
Our method struggles with such difficult cases, which are typically handled by eye-tracking techniques. 
A possible extension could be to include an eye-tracking branch.

\section{Social network analysis: \textit{Friends}-ness}
\label{sec:friends}

One principal way of signaling an interest in social interaction is the willingness of people to LAEO~\cite{goffman2008public,loeb1972mutual}. 
The duration and frequency of eye contact reflects the power relationships, the attraction or the antagonism between people~\cite{abele1986gaze}. 
Here, we present an application of our LAEO-Net to a social scenario: given head tracks we automatically infer relationships between people (\eg liking each other, romantic relationships) based on the frequency and duration that people LAEO over time.  
In particular, the idea behind \textit{Friends-ness} is to compute 
the ratio between the number of frames that people are LAEO over the frames that they share a scene. 
The higher the ratio, the more they interact.

\begin{figure}[t!]
\begin{center}
\includegraphics[width=0.90\linewidth]{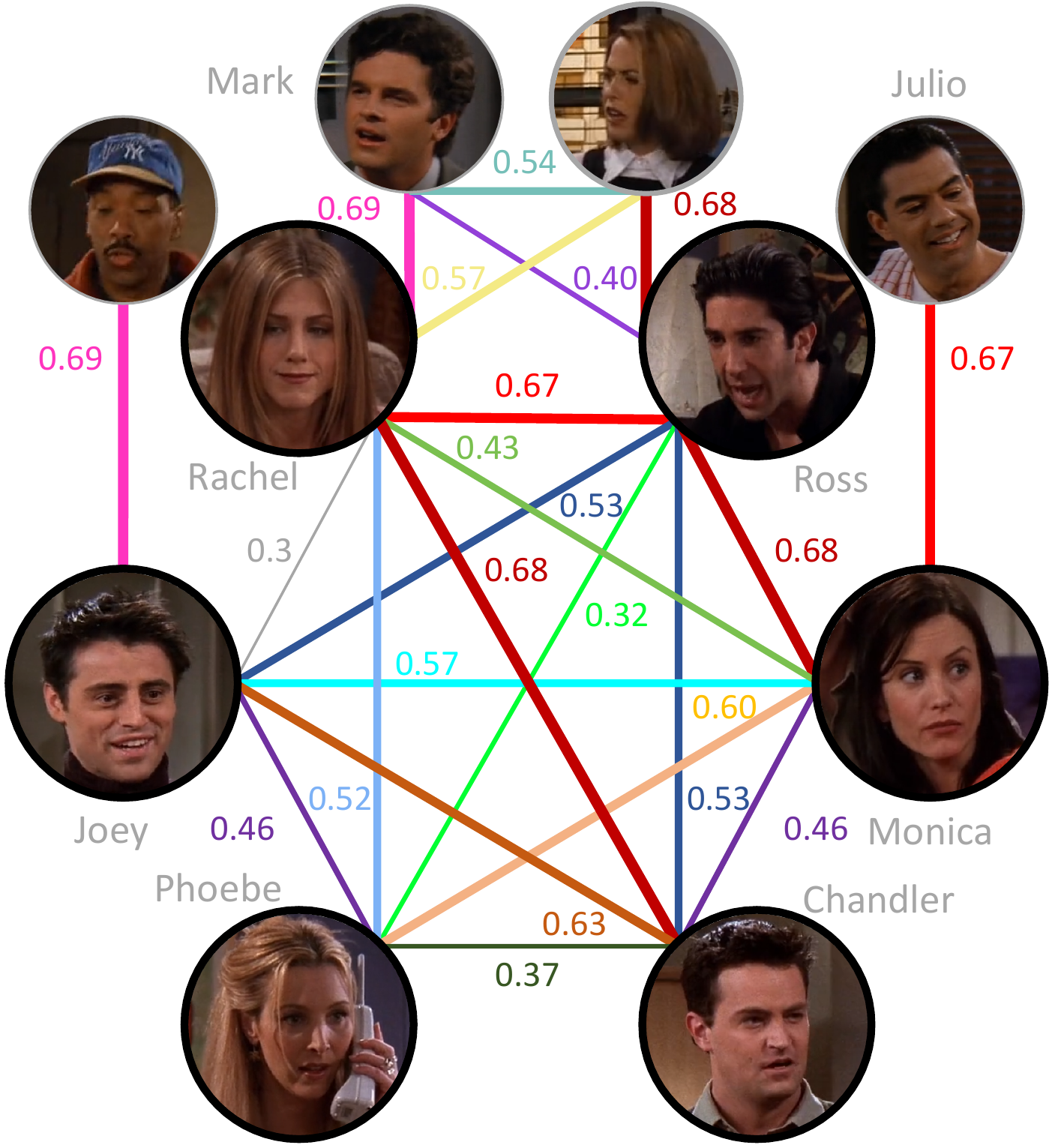}
\end{center}
\vspace{-4mm}
 \caption{\small{\textbf{\textit{Friends}-ness on Friends.} We measure `friends-ness' with the average LAEO score of each character pair and depict it with the edges in the graph: the thicker the edge, the more dominant the relationship. 
 Some patterns are clear: Ross and Rachel or Monica and Julio `like' each other more than Chandler and Phoebe or Ross and Mark. 
 (\textit{Best viewed in digital format.})
 }}
 \label{fig:rankfriends}
 \vspace{-4mm}
\end{figure}

\paragraph{Dataset and procedure. }
For this experiment, we use two episodes of the TV show `Friends' (\textit{s03ep12} and \textit{s05ep14}). 
First, we detect and track all heads, which results in almost $3k$ head tracks. 
Then, without any further training, we apply the LAEO-Net on each track pair to determine if two characters are LAEO. 

\paragraph{Character annotation.}
To determine which head track corresponds to which character, we annotate all produced head tracks as depicting one of the main characters 
(more than half of the tracks), irrelevant characters ($\sim30\%$), being wrong ($15\%$) or some secondary ones (the rest). 

\paragraph{Experiments and discussion. }
Our goal is to automatically understand underlying relationships between characters. 
Therefore, we measure the `likeness' between two characters as the average LAEO score over the frames two characters co-exist, and depict it in Figure~\ref{fig:rankfriends} (the thicker the edge, the higher the score and the stronger the relationship).  
We observe that the LAEO score captures the dominant relationships between characters, \eg Ross and Rachel vs characters that are more distant, \eg Phoebe and Chandler. 
Our study reveals all prominent pair relationships, demonstrating that the more people are LAEO, the stronger their \textit{interaction} and \textit{social relationship}.

\section{Conclusions} 
\label{sec:conclus}

In this paper, we focused on the problem of people \textit{looking at each other (LAEO)} in videos. 
We proposed LAEO-Net, a deep \textit{track} architecture, which takes as input head tracks and determines if the people in the track are LAEO. 
This is the first work that uses \textit{tracks} instead of bounding-boxes as input to reason about people on the whole track.  
LAEO-Net consists of three branches, one for each character's tracked head and one for the relative position of the two heads.  
Moreover, we introduced two new LAEO video datasets: \UCOLAEO and AVA-LAEO. 
Our experimental results showed the ability of LAEO-Net to correctly detect LAEO events and the temporal window where they happen. 
Our model achieves state-of-the-art results on the TVHID-LAEO dataset. 
Finally, we demonstrated the generality of our model by applying it to a social case scenario, where we automatically infer the social relationship between two people based on the frequency they LAEO.


\paragraph{Acknowledgements.}
We are grateful to our annotators: RF, RD, DK, DC and EP, and to NVIDIA for donating some of the GPUs used in this work. 
This work was supported by the Spanish grant ``Jos\'e Castillejo'', 
the EPSRC Programme Grant Seebibyte EP/M013774/1, and the Intelligence Advanced Research Projects Activity (IARPA) via Department of Interior/ Interior Business Center (DOI/IBC) contract number D17PC00341. 

\ifx
The U.S. Government is authorized to reproduce and distribute reprints for Governmental purposes notwithstanding any copyright annotation thereon. Disclaimer: The views and conclusions contained herein are those of the authors and should not be interpreted as necessarily representing the official policies or endorsements, either expressed or implied, of IARPA, DOI/IBC, or the U.S. Government.
\fi 

{\small
\bibliographystyle{ieee}
\bibliography{shortstrings,laeo.bib}
}

\end{document}